\documentclass[letterpaper, 10 pt, conference]{ieeeconf}  

\IEEEoverridecommandlockouts                              

\usepackage{graphics} 
\usepackage{xcolor,amsmath, graphicx, float}
\usepackage[breakable, theorems, skins]{tcolorbox}
\usepackage{multirow}
\usepackage{makecell}
\usepackage{booktabs}
\let\labelindent\relax
\usepackage[inline, shortlabels]{enumitem}
\usepackage{color}

\usepackage[hidelinks]{hyperref}

\usepackage{soul}
\newfloat{listing}{H}{lop}
\floatname{listing}{Listing}

\title{\LARGE \bf
Probing Multimodal LLMs as World Models for Driving
}

\author{\href{https://www.linkedin.com/in/shiva-sreeram-222b54250/}{ \color{violet}Shiva Sreeram$^1$},\quad \href{https://zswang666.github.io/}{\color{violet}Tsun-Hsuan Wang$^1$}, \quad
\href{https://alaamaalouf.github.io/}{\color{violet}Alaa Maalouf$^{1}$},\quad  
\href{https://www.tri.global/about-us/dr-guy-rosman}{\color{violet}Guy Rosman$^2$}, \\
\href{https://karaman.mit.edu/}{\color{violet}Sertac Karaman$^3$},  and 
\href{https://www.csail.mit.edu/person/daniela-rus}{\color{violet}Daniela Rus$^1$}\\ \\
{\color{magenta}$^1$MIT CSAIL} \quad\quad {\color{magenta} $^{2}$TRI } \quad\quad 
 {\color{magenta} $^{3}$MIT LIDS }\\ \\%
\href{https://github.com/sreeramsa/DriveSim.git}{\color{blue}GitHub Repository}
\thanks{
*This work is supported by Toyota Research Institute (TRI).  It, however, reflects solely the opinions and conclusions of its authors and not TRI or any other Toyota entity. {$^1$MIT CSAIL}, { $^{2}$TRI },  
 { $^{3}$MIT LIDS}.}
}

\newcommand{\METHOD}{\textsc{\texttt{DriveSim}}}

\newcommand{\DATA}{\textsc{\texttt{Eval-LLM-Drive}}}
\begin{document}

\maketitle
\thispagestyle{empty}
\pagestyle{empty}

\begin{abstract}
We provide a sober look at the application of Multimodal Large Language Models (MLLMs) in autonomous driving, challenging common assumptions about their ability to interpret dynamic driving scenarios. Despite advances in models like GPT-4o, their performance in complex driving environments remains largely unexplored. Our experimental study assesses various MLLMs as world models using in-car camera perspectives and reveals that while these models excel at interpreting individual images, they struggle to synthesize coherent narratives across frames, leading to considerable inaccuracies in understanding (i) ego vehicle dynamics, (ii) interactions with other road actors, (iii) trajectory planning, and (iv) open-set scene reasoning. We introduce the \DATA{} dataset and \METHOD{} simulator to enhance our evaluation, highlighting gaps in current MLLM capabilities and the need for improved models in dynamic real-world environments.
\end{abstract}



\section{INTRODUCTION}
In the rapidly evolving field of artificial intelligence, Multimodal Large Language Models (MLLMs), such as GPT-4o \cite{gpt4o}, have demonstrated unprecedented capabilities in understanding and generating image/text-based content~\cite{LLMSAPPSSURVEY}. 
Recently, MLLMs have been introduced to the realms of driving to improve context understanding ~\cite{chen2023driving}, extract spatial features from frames to teach a driving policy based on said features, improving the generalization ability of autonomous driving~\cite{wang2024drive}, infer system requirements from in-cabin users' commands to meet their intent ~\cite{Yang2024WACV}, understand the driving environment~\cite{fu2024drive}, and more~\cite{cui2024survey}. 
However, the performance of these powerful models has not been tested for \textbf{scene} (sequence of images) reasoning in a dynamic driving context, let alone one in a closed control loop, and thus, remains an intriguing area of exploration. We ask the question: 
\begin{itemize}
    \item {\textit{``\ul{Can MLLMs operate as driving world models}''?}}
\end{itemize}

\noindent\textbf{Our contribution. } To this end, in this work, we study the reasoning capabilities of MLLMs within driving scenarios, aiming to measure their applicability in understanding complex, \textit{dynamic} environments in a variety of scenarios, and their ability to take appropriate actions in decision-making through the integration of \textit{a sequence} of visual data captured from a fixed camera mounted on a driving car \textit{as if the MLLM was the driver}. Specifically, we offer:
\begin{itemize}[align=right,itemindent=0em,labelsep=2pt,labelwidth=1em,leftmargin=*,itemsep=0em]
\item  \label{firstcont} A comprehensive experimental study to evaluate leading MLLMs in their ability to understand and make decisions in dynamic driving scenarios, involving both real driving footage and closed-loop controlled driving. 
Our tests cover multiple facets of environmental interactions: ego-car dynamics, interactions with other road actors, trajectory planning, and open-set driving scene reasoning. Surprisingly, our findings reveal that MLLMs struggle with interpreting/reasoning and taking correct actions in dynamic driving scenes with significant inaccuracies and biases. 
\item ``\DATA'': A new dataset designed to provide an array of driving scenarios for evaluating the capabilities of MLLMs in understanding and reasoning about real-world driving scenes from a fixed in-car camera perspective, the same as the driver viewpoint. 
Real footage captured on the road is the basis of this data, alongside a closed-loop controlled driving simulator \METHOD{} to generate additional diversity and open-set scenarios to the dataset. 
\end{itemize}

\begin{figure}[t]
  \centering
  \includegraphics[width=1\linewidth]{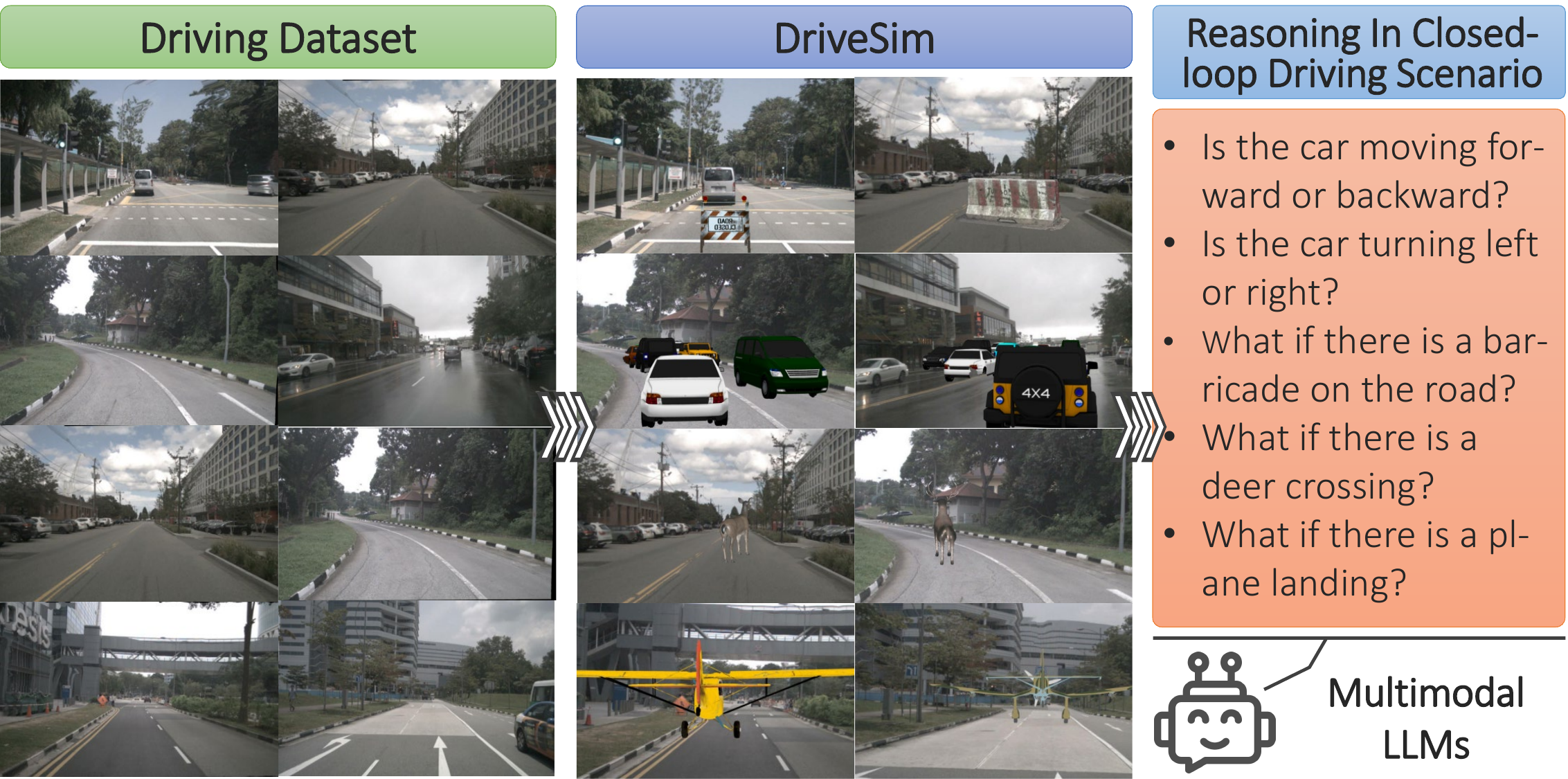}
  \label{fig:teaser}
  \caption{\textbf{Are MLLMs world models for driving?} 
  We investigate their effectiveness in understanding and reasoning about dynamic driving scenarios from sequential images with an introduced real-world driving and re-simulated dataset. Our experiments show that MLLMs struggle to form coherent narratives, failing to reason about car motion, traffic, etc.
  } 
\end{figure}

\begin{figure*}[t]
  \centering
  \includegraphics[width=1\linewidth]{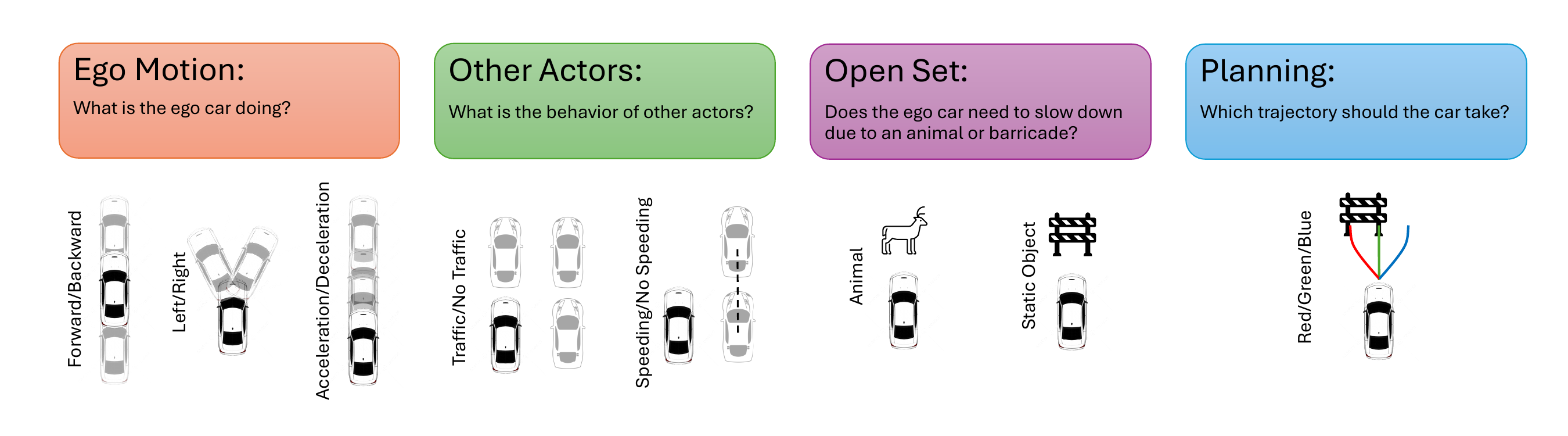}
  \caption{Components of what a model must understand to be a world model for driving.}
  \label{fig:components-of-world-model}
\end{figure*}
\noindent\textbf{A glimpse into our findings. }Our experimental results, 
reveal a paradox in the performance of MLLMs. While these models excel at understanding individual images, they struggle to synthesize a coherent narrative across sequences depicting dynamic behavior.
This is especially evident in their difficulty in reasoning about vehicular motions, such as identifying whether the ego-car is moving forward or backward. This is headlined by the fact that GPT-4V  predicted all scenes as forward-moving, a trend seen in 78\% of cases with GPT-4o as well! This may stem from a bias in training data, where vehicles predominantly move forward. In planning, these models and others (including Claude3, LLaVA-1.6, InstructBLIP, and more) consistently failed.
When focusing on ChatGPT, it is clear that improvements have been made since the legacy of the GPT-4V model to the latest GPT-4o model, particularly in identifying the dynamic interplay of other road actors. However, there are still failures in other aspects, such as in ego-car dynamics, that prevent it from holding the status as a driving world model. While this shows a positive trajectory in the development of these models, the results highlight additional areas of improvement for the top models.
In summary, the experiments highlight a critical gap in the models' ability to connect discrete visual information over time to infer motion, suggesting a limitation in their current state when it comes to understanding the fluidity and continuity inherent in real-world dynamics. 


\begin{table*}[t]
    \centering
    \small
    \caption{Overall accuracy. Evaluating MLLM performance in driving tasks requiring reasoning.}
    \scalebox{0.74}{
    \begin{tabular}{c || c c c| c c c| c c c | c c c |c c c | c c | c}
    \toprule
    \multirow{3}{*}{Model} & \multicolumn{9}{c|}{ {Ego Motion} } & \multicolumn{6}{c|}{Other Vehicles} & \multicolumn{2}{c|}{Open-set Reasoning } & \multirow{2}{*}{Planning} \\
     & \multicolumn{3}{c}{Forward/Backward} & \multicolumn{3}{c}{Accelerate/Decelerate}  & \multicolumn{3}{c|}{Left/Right} & \multicolumn{3}{c}{Speeding} & \multicolumn{3}{c|}{Traffic} & Object/Animal & Plane Landing &  \\
    & Real & Sim & Both & Real & Sim & Both & Real & Sim & Both & Real & Sim & Both & Real & Sim & Both &  Sim & Sim & Sim \multirow{3}{*}{}\\
    \midrule
    MiniGPT4-v2 & 0.50 & 0.50 & 0.50 & 0.50 & 0.50 & 0.50 & 0.55 & 0.42 & 0.48 & 0.50 & 0.50 & 0.50 & 0.50 & 0.50 & 0.50 & 0.20 & 0.67 & 0.30 \\
    InstructBLIP & 0.50 & 0.50 & 0.50 & 0.43 & 0.47 & 0.45 & 0.50 & 0.50 & 0.50 & 0.50 & 0.50 & 0.50 & 0.57 & \hl{\textbf{0.73}} & 0.65 & 0.25 & 0.50 & 0.30 \\
    LLaVA-1.6 & 0.48 & 0.47 & 0.48 & 0.50 & 0.47 & 0.48 & 0.47 & 0.53 & 0.50 & 0.50 & 0.50 & 0.50 & 0.68 & 0.53 & 0.61 & 0.27 & 0.50 & 0.25 \\
    GPT-4V & 0.50 & 0.50 & 0.50 &\hl{\textbf{0.57}} & \hl{\textbf{0.55}} & \hl{\textbf{0.56}} & 0.60 & 0.48 & 0.54 & 0.52 & 0.55 & 0.53 & 0.63 & 0.62 & 0.63 & \hl{\textbf{0.80}} & 0.63 & 0.40 \\
    Claude3 & 0.55 & 0.48 & 0.52 & 0.48 & 0.47 & 0.48 & 0.52 & 0.50 & 0.51 & 0.50 & 0.55 & 0.53 & 0.43 & 0.62 & 0.53 & 0.70 & 0.53 & \hl{\textbf{0.45}} \\
    GPT-4o & \hl{\textbf{0.60}} & \hl{\textbf{0.52}} & \hl{\textbf{0.56}} & 0.55 & 0.40 & 0.48 & \hl{\textbf{0.62}} & \hl{\textbf{0.65}} & \hl{\textbf{0.63}} & \hl{\textbf{0.72}} & \hl{\textbf{0.77}} & \hl{\textbf{0.74 }}& \hl{\textbf{0.80}} & \hl{\textbf{0.73}} & \hl{\textbf{0.77}} & 0.73 & \hl{\textbf{0.70}} & \hl{\textbf{0.45}} \\
    \bottomrule
    \end{tabular}}
    \label{tab:accuracy}
\end{table*}

\section{RELATED WORK}
Lately, the move toward combining different modalities into single large-scale models has gained momentum, such as CLIP~\cite{pmlr-v139-radford21a}, BLIP~\cite{li2023blip}, GPT-4V~\cite{achiam2023gpt} and others~\cite{alayrac2022flamingo}.

\noindent\textbf{MLLMs in robotics}. Recent advancements in robotics have integrated MLLMs, demonstrating their proficiency in engaging effectively within dynamic open-set environments, such as for constructing 3D maps \cite{ding2023pla,huang2023audio}, in control and planning \cite{li2022pre,brohan2022rt,bisk2020experience,tellex2020robots,ahn2022can}, 
in understanding 3D scenes \cite{peng2023openscene,jatavallabhula2023conceptfusion}, and in systems for detection and tracking \cite{maalouf2024follow,li2022language,ghiasi2022scaling,liu2023grounding}.
Additionally, these models have shown broad adaptability over multi-modal data \cite{ramesh2022hierarchical,maalouf2023follow,jatavallabhula2023conceptfusion,ramesh2021zero,patashnik2021styleclip,crowson2022vqgan}, 
heralding a new phase where robots can make wise decisions and interact with their surroundings.

\noindent\textbf{MLLMs for driving.} 
In driving, explainable and language-driven representations have gained attention for introspection and event analysis \cite{Kim2019-fw,tan2023language,zhong2023language,omeiza2021explanations,kuo2022trajectory}. Integrating MLLMs into autonomous vehicles enhances vehicle intelligence and user interaction \cite{cui2024drive} by leveraging real-time data (e.g., traffic, weather) to improve awareness \cite{cui2023receive} and navigation \cite{sriram2019talk}. LLMs facilitate user-friendly communication for planning \cite{wang2024describe,omama2023alt,NEURIPS20236b8dfb8c} and personalize driving settings \cite{sha2023languagempc}. They also improve generalization and explainability \cite{wang2024drive}, enhance context awareness \cite{chen2023driving}, interpret user commands \cite{Yang2024WACV}, and better understand driving environments  \cite{fu2024drive}.

\noindent\textbf{Simulation in driving. }
Training and evaluation of robotic controllers via simulation have become a dominant approach, as evidenced by~\cite{tedrake2019drake,dosovitskiy2017carla,shah2018airsim},
specifically, in driving~\cite{amini2022vista, caesar2020nuscenes,Lu2024}. However, even these simulated environments (i) can not fully encapsulate the range of vehicle dynamics (e.g., moving forward or backward, acceleration or deceleration, turning left or right, and more), (ii) lack support for adding dynamic characters to scenes for generating interesting driving behaviors (e.g., speeding cars and traffic), and (iii) most importantly they are not labeled, making it hard to use for evaluating MLLMs as world models for driving. 
\section{PROBING FROM A DATA PERSPECTIVE}
\label{sec:setup}

Ultimately, 
a driving world model should encompass multiple facets of environmental interactions and scene reasoning, given in Fig.~\ref{fig:components-of-world-model}, which we define and test as follows: 
\begin{enumerate*}
[(i),itemsep=0pt, topsep=0pt, partopsep=0pt, parsep=0pt]
\item \textbf{Ego-car dynamics: } We check the models' ability to grasp fundamental driving dynamics, such as directionality (forward/backward), velocity changes (acceleration/deceleration), and road adjustments (turning right/left), requiring an understanding in geometric and temporal aspects.\label{eval1} 
\item
\textbf{Dynamic interplay of other road actors: }Progressing beyond the basics, we then challenge the models to reason about the dynamic interplay of other road actors: detecting fast-moving vehicles and discerning traffic jams. \label{eval2}  
\item 
\textbf{Planning ability: }Then, we examined the ability of the models to plan accurate driving trajectories, checking whether they can effectively reason the means to avoid obstacles along the way. \label{eval3} 
\item \label{eval4} \textbf{Open-set scene reasoning: }
The true test of adaptability lies in open-set reasoning, where our testing challenges conventional driving expectations by creating unforeseen scenarios
as unpredictable as airplanes landing on roads or sudden animal appearances, pushing the boundaries of what MLLMs can anticipate and interact correctly with in the meticulously crafted world model.
\end{enumerate*}
These multi-layer testing scenarios challenge the models' interpretability and decision-making, offering insights into MLLMs' contributions to real-world applications, from alerting wrong-ego car behavior to enhancing navigation systems with real-time alerts on strange (open-set) scenes, to traffic updates, and to driving validation and planning.




\subsection{Providing the Means to Evaluate a Driving World Model}
Surprisingly, the evaluation of MLLMs' scene reasoning in the context of dynamic driving scenarios in the closed-loop control setting remained largely unexplored, potentially due to the lack of a suitable dataset or simulator. 
To address this gap, we introduce \DATA; a dataset designed to test MLLMs as driving world models, focusing on the components of a world model and evaluations \ref{eval1}--\ref{eval4}. 
The dataset is divided into two parts: (i) real road footage capturing vehicle dynamics and interactions with other road users, valuable for testing MLLMs reasoning in real scenes, and (ii) scenes generated by \METHOD{} using closed-loop sensor synthesis. 
We can expand upon the real footage with additional diversity through direct manipulation of ego vehicle dynamics and other actors.  Furthermore, the simulator enables the generation of various open-set scenes involving characters like animals, barriers, and vehicles that are not feasible to obtain in everyday driving conditions, yet enhance the platform's utility for probing models.
\subsection{Scenarios directly from the real world}

We drove a 2019 Lexus RX 450H with 
a 30Hz BFS-PGE-23S3C-CS RGB camera to collect data in Cambridge, MA, and the surrounding area. For most cases, the footage consists of 50\% daytime footage, 30\% evening/dusk, and 20\% night. 
This was all collected in a manner that ensured following rules-of-the-road and safe driving practices, described as follows. 
\textbf{Acceleration/deceleration} captured scenarios involve the ego car legally speeding up (e.g., entering a highway) or slowing down (e.g., approaching a stop sign or turn).
\textbf{Turns} include both standard intersection turns and subtle curves angling left or right.
\textbf{Forward} videos show standard driving ahead. Due to traffic rules, \textbf{backward} videos are created by reversing forward footage that contains no other moving actors like pedestrians or vehicles.
\textbf{Traffic} is denoted as situations where traffic causes the ego vehicle to slow down, notably not due to a traffic light but from natural congestion in the road. Since nighttime traffic is less common, this data has a higher proportion of daytime footage (70\% day, 30\% evening) compared to other splits.
The real footage does not involve as bumper-to-bumper traffic as we later show can be generated by the simulator due to appropriately safe stopping distances, but still requires the ego vehicle to slow down. 
Finally, a \textbf{``speeding'' vehicle} is present when another vehicle overtakes the ego vehicle, and through prompting, the ego vehicle is clarified to be driving at the speed limit, so the overtaking vehicle is implied to be speeding.

\subsection{Scenarios by re-simulation of real-world data}

\noindent\textbf{Closed-loop sensor synthesis and control. }
To meet the requirements of our experimental setup, which necessitates a controlled environment and counterfactual testing (as in generating counterfactual data different from the original dataset as opposed to counterfactual reasoning of MLLMs), we develop a data-driven simulator on top of the nuScenes dataset \cite{caesar2020nuscenes}. This approach effectively balances sensor realism \cite{chen2021geosim,yang2023unisim}, closed-loop simulation \cite{suo2021trafficsim,amini2022vista}, and scenario setup controllability \cite{li2024scenarionet,ding2023realgen}, making it an ideal match for our use case. In the subsequent sections, we outline the key features of the simulator and elucidate their significance to our empirical study for comprehending the reasoning processes of MLLMs within driving scenarios.

\noindent\textbf{Object and actor synthesis in the scene. }
Building on the described 3D reconstruction pipeline, we seamlessly integrate 3D meshes of desired objects and actors into the scene. These meshes can be efficiently sourced from Objaverse dataset~\cite{deitke2023objaverse} by leveraging the textual comprehension abilities of LLMs on their annotations. For instance, we can identify annotations suggesting that the corresponding meshes represent animals.
Utilizing the map's geometric and semantic information, we strategically position the meshes in plausible locations and orientations. Examples include beside the same lane as the ego car, beneath the traffic light, etc.

\noindent\textbf{Behavior modeling of actors. }
For the behavior of ground vehicles, we employ a Proportional-Integral-Derivative (PID) controller \cite{ang2005pid} for steering control to track a reference path derived from either the map or a motion plan; for acceleration control, we use an Intelligent Driver Model (IDM) \cite{treiber2000congested} focused on the nearest actor ahead of the ego car moving in a direction potentially in collision with the ego car.
For motion planning, we deploy a state lattice planner with quintic polynomial trajectory generation \cite{howard2008state}, in which the target state lattice is determined to be a specific distance ahead of the ego car in the local frame of its current lane or adjacent lanes.
For behavior modeling of other actors, we create trajectories through spline interpolation from predefined start to end poses.
Our focus is on modeling the behavior of synthetic actors in reaction to the ego car, to themselves, and to other pre-existing actors or objects in the scene, rather than the behaviors of those already existing entities.

\section{EXPERIMENTAL STUDY}
\begin{figure*}[t]
    \centering
    \includegraphics[width=\linewidth]{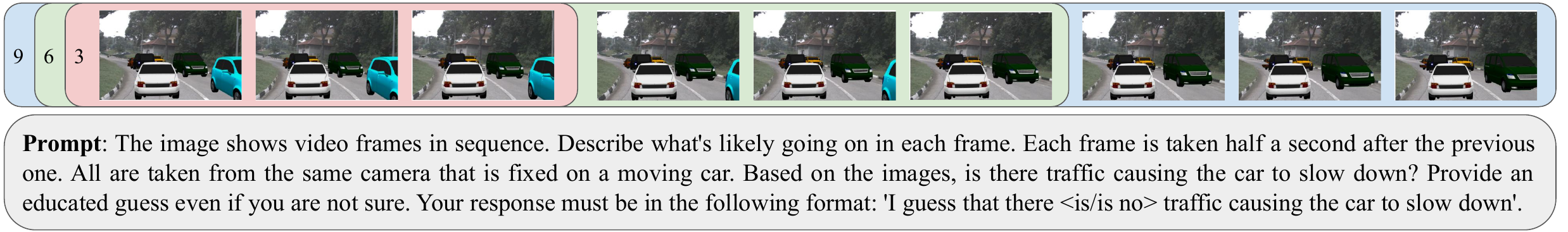}
    \caption{Heavy traffic scene provided by \METHOD{} converted into a grid alongside the text prompt.}
    \label{fig:vid-setup}
\end{figure*}

\noindent\textbf{Methodology.} We used the paradigm explained in Section~\ref{sec:setup} to test SOTA MLLMs' abilities to determine ego-car motion: (1) is the car proceeding forward or backward? (2) is it accelerating or decelerating? (3) is it turning left or right? All in a categorical manner. 
Then, we evaluate their reasoning capabilities on other factors in the street to determine whether it detects a speeding car; (4) is there a speeding car?, or heavy traffic; (5) is there heavy traffic?
Additionally, we test the decision-making of MLLMs based on an open-set (or even weird) environment by generating open-set scenes by \METHOD{}, such as providing images with the sudden appearance of an animal or static object and even a plane landing: (6) can the ego-car keep moving in the same lane?. 
We finally can test the capabilities of MLLMs to pick the best trajectory in navigating around obstacles while trying to remain in the lane: (7) which trajectory is the best to follow? 


\noindent\textbf{Representing a video scene. }
We aimed to provide video input to the models to replicate the camera view in real-world driving scenarios. 
From video input, we create a grid~\cite{fusseldieb2023readvids} of video frames where each frame is half a second apart, and of resolution $399 \times 224$. We test a varied number of frames: three, six, and nine. The total grid resolution is $1259 \times 244$/$599$/$712$ for grids of three, six, and nine frames respectively, including all frames and white spaces to split them.  
This format was utilized to avoid concerns with models parsing the images in multi-query approaches and avoiding context length limits while providing a high number of frames. This setup is showcased in Fig.~\ref{fig:vid-setup}.

\noindent\textbf{Dataset. } For each question from (1)--(5), we experiment with two sets, the first set is of real, on-the-road footage collected by the authors, and the second set is generated by \METHOD{}. For question (6) we rely on \METHOD{} to generate (rare/weird) open set scenes that are hard to capture in real streets. For each of these questions ((1)--(6)), we collect 20 videos in simulation; since each video is parsed in a grid of $3$, $6$, and $9$, the total number of videos of a question is $60$ in simulation and an additional $60$ in real footage (similarly from 20 videos) for questions (1)--(5). 
For question (7), we use \METHOD{} to generate trajectory choices over the road surface and static objects, in five scenes and four scenarios per scene, leading to $20$ datapoints.

\noindent\textbf{Prompting. }Alongside frames, we must provide an appropriate prompt which informs the model of the format of the image, that the frames come from a camera fixed on a moving car, and obtain a response to the relevant question. This is shown in an example prompt in Fig.~\ref{fig:vid-setup} which queries whether the ego vehicle is experiencing traffic or no traffic. We follow a similar format when prompting for ego actions and the other actors behavior scenario. We ask to describe what is likely going on in each frame to guarantee the model understands it is parsing a video and in the correct frame order, we can then manually validate its explanation. Additional components asked the model to ``guess'' and answer ``in the following format'' to prevent scenarios where the model reports that it is not confident enough to tell what the answer is. The prompts' phrasing was carefully formulated after experimentation with the models' responses, to ensure that the models accurately processed the frames in the correct sequence order, recognized that the footage was from a camera fixed on a moving car, and provided clear, relevant responses to the questions.
\begin{figure}[t]
    \centering
        \includegraphics[width=\linewidth]{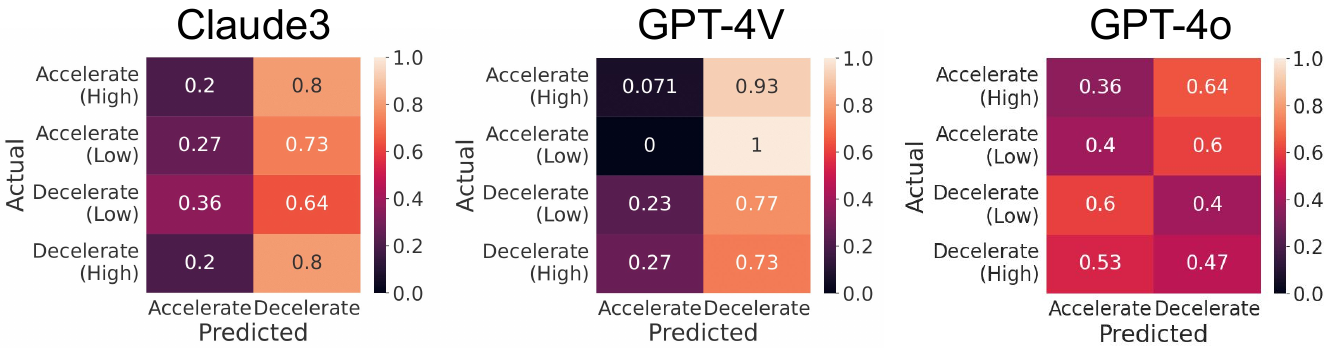}
        \caption{Accelerate vs decelerate: Confusion matrices.}
        \label{fig:acc-plot}
\end{figure}

\noindent\textbf{Evaluation and reported results. } We compare the obtained results from the MLLM to the ground truth created by the authors. Table \ref{tab:accuracy} reports how GPT-4o, Claude3, GPT-4V (being GPT-4 Legacy model as of Sep. 2024), LLaVA-1.6, InstructBLIP, and MiniGPT4-v2 perform in these cases. To further expand on the results of this evaluation process given in Table \ref{tab:accuracy}, we delve into specifics for ego-motion, other actor behavior, open-set, and planning reasoning. We provide a more sophisticated analysis focusing on GPT-4o, GPT-4V (a snapshot of which from March 2024), and Claude3 due to their nature as some of the largest models available and the higher levels of reasoning observed in our evaluation, in addition to tracking the improvements of the top of the line GPT-4 model from March compared to September 2024.

\subsection{Ego Motion Reasoning}
\label{sec:ego_reasoning}
\begin{figure*}[t]
  \centering
  \includegraphics[width=0.325\textwidth]{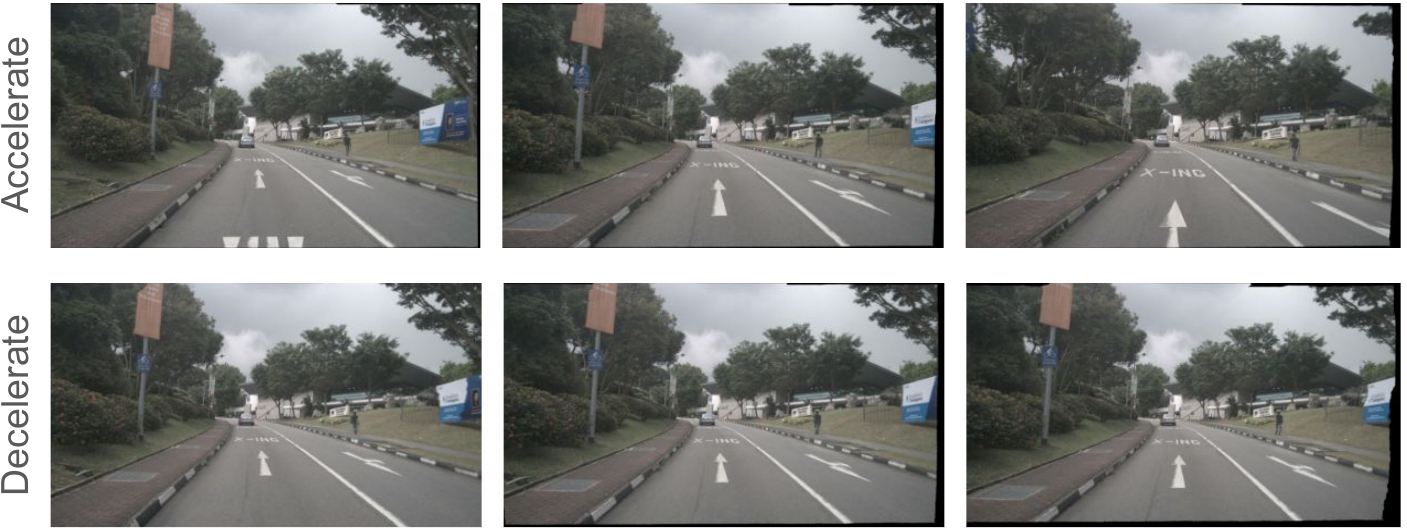}
  \includegraphics[width=0.3\textwidth]{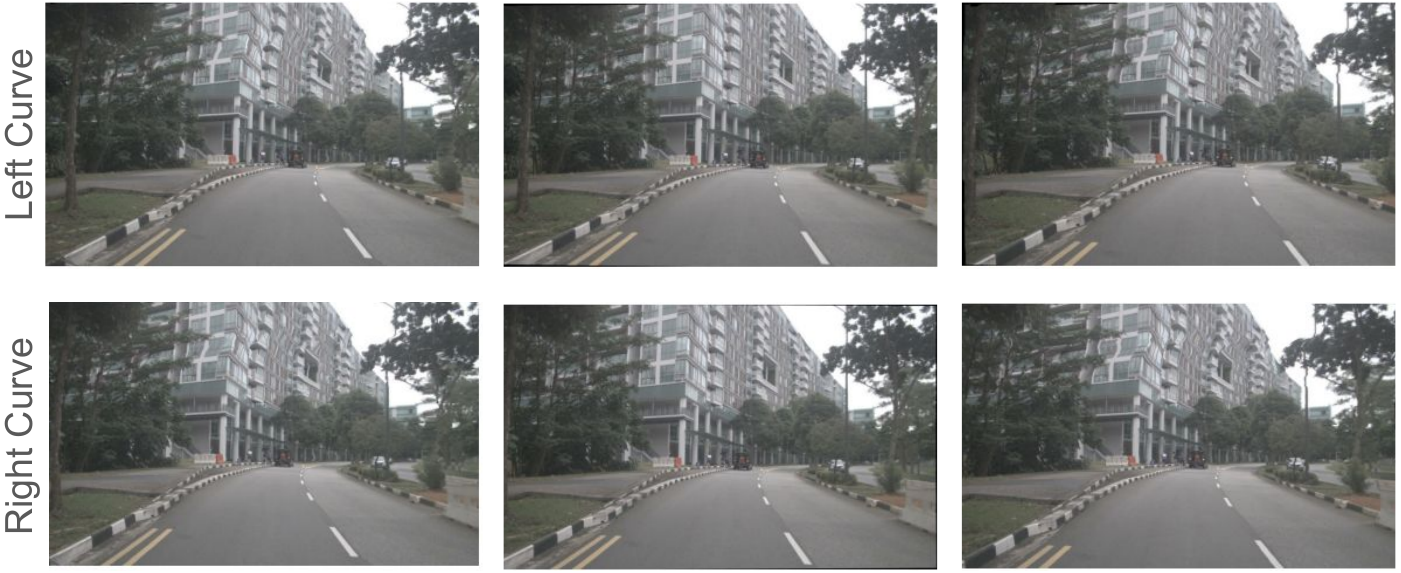}
  \includegraphics[width=0.34\textwidth]{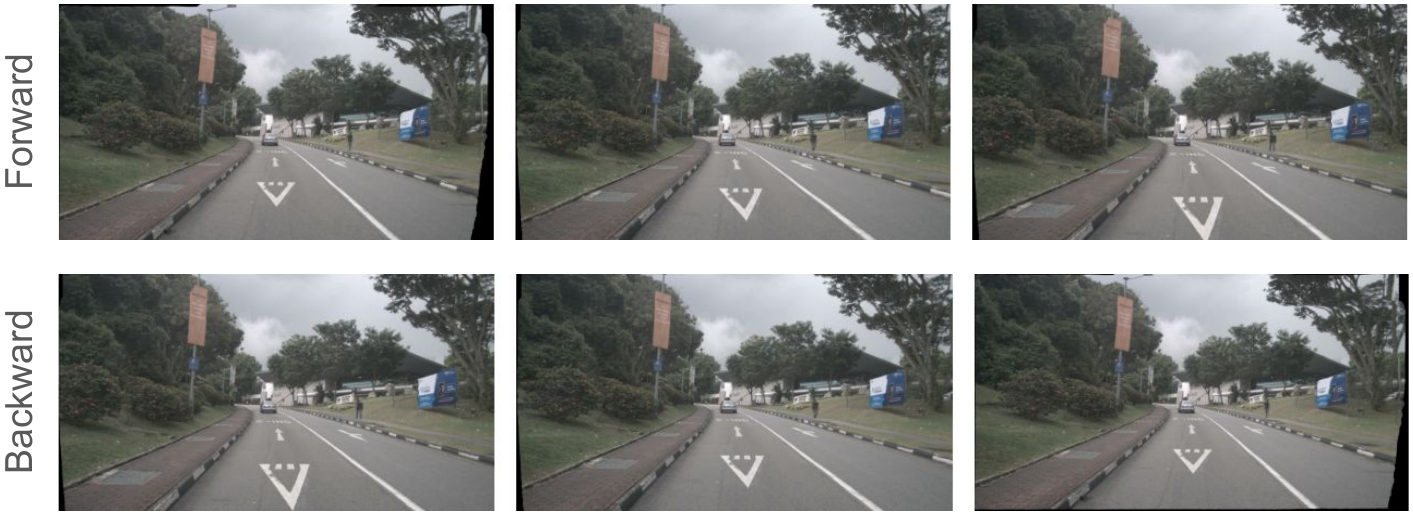}
  \caption{Frames of ego-motion videos of scenes provided by \METHOD.}
  \label{fig:ego-motion-frames}
\end{figure*}

\noindent\textbf{Acceleration vs deceleration.} 
In simulator datasets, \METHOD{} generates scenarios where we traverse through the same scene at different rates of acceleration or deceleration, both at high and low rates. The human eye can determine \textit{acceleration}/\textit{deceleration} by using a reference point in view and observe the change in distance over time (such as the arrow marking on the road in Fig.~\ref{fig:ego-motion-frames}), so let us explore the models' capability to do so.
As shown in Table~\ref{tab:accuracy}, for most models the performance was roughly 50\% across all data, if not worse. 
To further understand the behavior, we explore the performance of Claude3, a snapshot of GPT-4V from March 2024, and GPT-4o in Fig.~\ref{fig:acc-plot} on the synthesized data as it provides us the flexibility to generate and verify several acceleration/deceleration rates. We present the confusion matrices for scenarios of high and low acceleration and deceleration rates to the models. Note that in these tests, GPT-4V (March) was biased towards a response of \textit{decelerate}. Interestingly, there were more cases where the model predicted \textit{accelerate} when the ground truth was \textit{decelerate} compared to when the ground truth was \textit{accelerate}. For Claude3, while the bias towards \textit{decelerate} remained, it was less extreme than for GPT-4V. In tracking the improvements from GPT-4V to -4o, we see that this bias is largely removed despite the $48\%$ accuracy overall.

\begin{figure}[t]
    \centering
    \includegraphics[width=\linewidth]{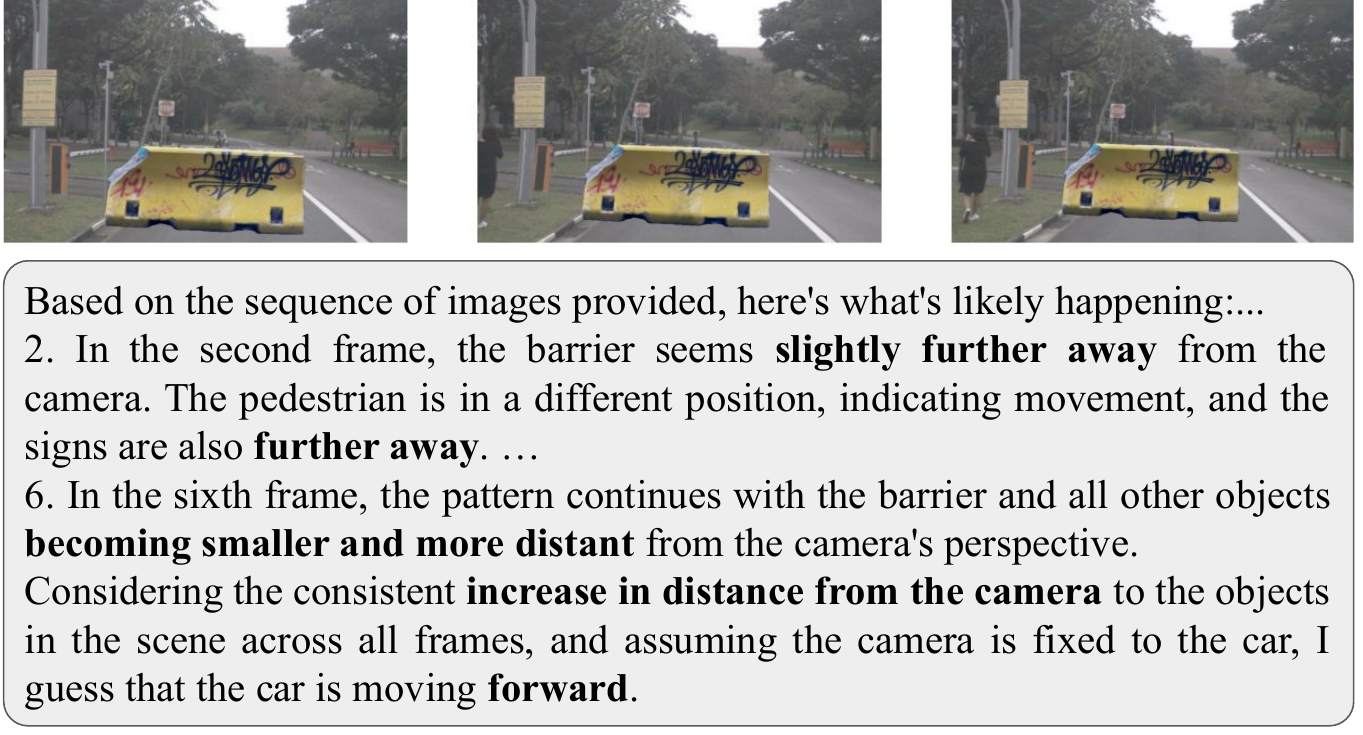}
    \caption{Frames of backward motion, with an added object, provided by \METHOD{}, alongside truncated GPT-4V response in analyzing the video.}
    \label{fig:special-back-frames}
\end{figure}

Overall, these results demonstrate a past bias in MLLMs where they frequently responded with decelerating. Although GPT-4o has largely eliminated this bias, its overall accuracy decreased, so the results still reveal limitations in MLLMs' ability to accurately reason whether the ego vehicle is accelerating, a crucial aspect of driving dynamics.

\noindent\textbf{Left vs right.} 
For the simulated set we particularly stress the models by evaluating the inputs while the ego vehicle curves towards the left or right while progressing at a constant speed, as opposed to turning into a perpendicular lane covered in real-driving footage. We test with high and low levels of curvature. The human eye can determine turning direction by the perspective shift of static elements of the scene, moving from the center to the edge of view over time. In the case of Fig.~\ref{fig:ego-motion-frames}, we see the trunk of different trees shift out of view depending on the turn. Let us explore how well the model can observe this.


\begin{figure}[t]
        \includegraphics[width=\linewidth]{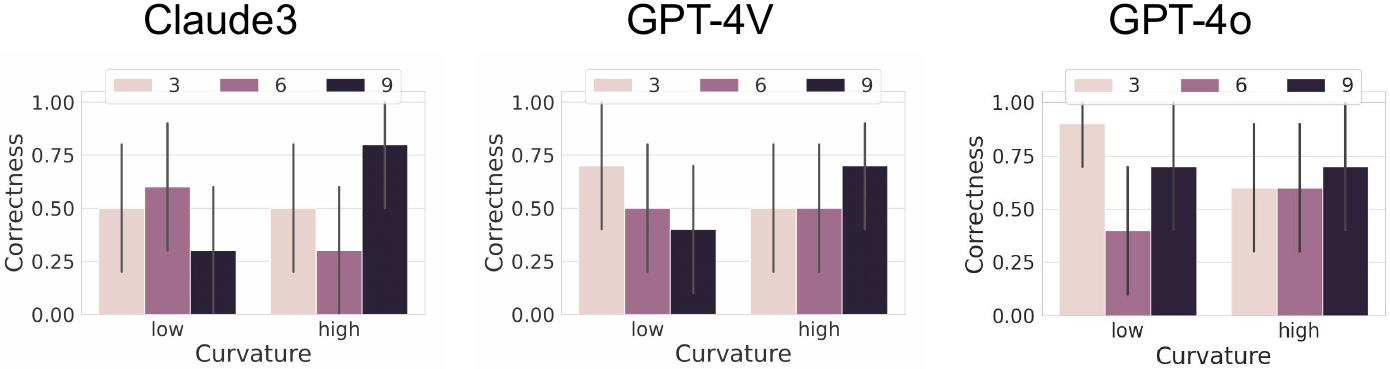}
        \caption{Left vs right: Performance bar plots.}
        \label{fig:leftright-plot}
\end{figure}

As shown in Table~\ref{tab:accuracy}, the results, again, were roughly 50\% for all models. Note that with the real data, the performance for some models was higher than simulation, but given that much of the data in the real data includes a turn into a perpendicular lane as opposed to curving along the road, this difference is expected. 
However, with control of the curvature in \METHOD{}, we can further analyze the performance of GPT-4V (March), GPT-4o, and Claude3 in Fig.~\ref{fig:leftright-plot}. While no clear bias was witnessed, we noticed a curious result when comparing the correctness of different levels of curvature for different numbers of frames provided to the model. For GPT-4V, we see that with low curvature, a lower number of frames yielded better results while for a high curvature, more frames led to a higher accuracy. While Claude3 does not have a clear trend, the model achieved the highest success out of all buckets when provided with a higher curvature and nine frames and the lowest success with a low curvature and nine frames. This behavior appears to indicate that with higher rates of curvature, more temporal information helps the model observe the differences in perspective shift, but if these differences are more subtle with low curvature, more temporal information harms the models' performance. This trend is not as strong in GPT-4o, with the six frames case in low curvature being especially different.

\begin{table}[t]
    \centering
    \scriptsize
     \caption{Percentage of \textit{forward} / \textit{backward} responses.}
     \scalebox{0.85}{
    \begin{tabular}{||c c c c c c c||}
     \hline
      & MiniGPT4 & InstructBLIP & LLaVA & GPT-4V & Claude 3 & GPT-4o \\ [0.5ex] 
     \hline\hline
     Forward & 100\% & 100\% & 96.7\% & 100\% & 86.7\% & 75.8\% \\ 
     \hline
     Backward & 0\% & 0\% & 2.5\% &  0\% & 13.3\% & 24.2\%\\ [1ex] 
     \hline
    \end{tabular}}
   
    \label{tab:forwardbackward}
\end{table}

\begin{figure}[t]
    \centering
        \centering
        \includegraphics[width=\linewidth]{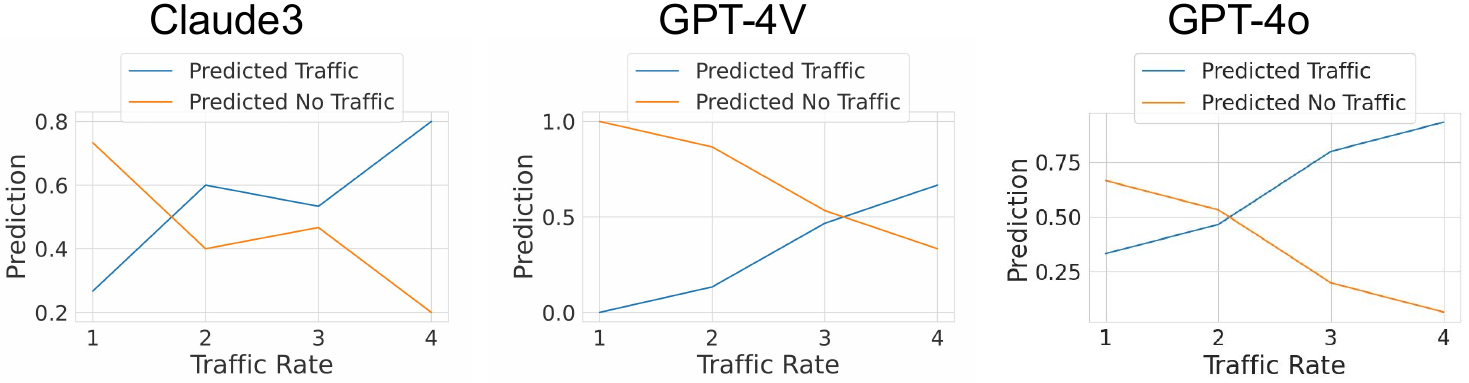}
        \caption{Line plots depicting a confusion matrix for model performance in simulated traffic vs no traffic scenes.}
        \label{fig:traffic-plot}
\end{figure}
\noindent\textbf{Forward vs. backward} motion comparison is intuitive for humans, relying on basic geometric and dynamic understanding (e.g., are static objects getting closer or farther such as the road markings in Fig.~\ref{fig:ego-motion-frames}?). This test examines models' geometric reasoning, using simulated data at varying constant speeds, similar to other ego-motion experiments.
We evaluate the models' understanding in this case.
Table \ref{tab:accuracy} shows the results here were once again roughly 50\% for all models. However, there is an extreme underlying bias in the model responses. We further analyze the performance in Table~\ref{tab:forwardbackward}. This table showcases the percentage of responses from the models that were either \textit{forward} or \textit{backward} regardless of the ground truth. Despite the differences in the scenes, 
GPT-4V (both the March 2024 snapshot and the Legacy model as of Sep. 2024), alongside most other models, \textbf{always reported that the ego vehicle is progressing forward}, with both Claude3 and GPT-4o having a strong bias to reporting forward. 
The bias here is at a level where it overwhelms the evidence and the model responds that if a car is on the road, it must be driving forward as technically, that is what the car should be doing. As such, using the capabilities of \METHOD, we further probe to understand the source of the model failure. Intending to try to get a \textit{backward} response from GPT-4V, we placed a notable object, a barrier, in the scene as the ego car drives back from it. In the three frames of this sequence shown in Fig.~\ref{fig:special-back-frames}, the barrier is getting further and further from the camera. 
Re-running the experiment with this video, we obtain a fascinating result (Fig.~\ref{fig:special-back-frames}). As stated, \textbf{the barrier is getting further} from the camera, a trait \textbf{the model is able to identify} as it parses the frames. \textbf{However,} even with this reasoning, \textbf{it predicts the ego vehicle is going forward}. 
The model has the capability of reasoning required to understand the geometry of the world, yet its responses are biased to such a level that it fails to make accurate predictions.


\subsection{Other Actor Behavior Reasoning}
\label{sec:other_reasoning}
\noindent\textbf{Traffic vs no traffic.} 
There are two main sources of traffic: the amount of other vehicles on the road and the speed at which the ego vehicle can move given the other vehicles. As such, the geometrical understanding is necessary to witness the number of other vehicles in the scene and the combination of geometric and temporal reasoning for the speed of traffic flow. In real-footage, we have no control over the other vehicles, so traffic is a very binary state, but with our simulator, we provide four levels of traffic. (i) The lowest level, with the label of \textit{no traffic}, is where there is no other vehicle in the same lane. (ii)  The second lowest level, also labeled \textit{no traffic}, is defined by another vehicle being in the same lane but moving at a high enough speed such that the ego's speed is not hindered. (iii) The next level, which we label as \textit{traffic}, is defined by a large number of other vehicles, with slow yet steady traffic flow. (iv) The highest, which we label as \textit{traffic}, involves a large number of actors all moving at a very slow speed, shown in Fig.~\ref{fig:vid-setup} where to the human eye, the level of traffic is clear.

To clarify the concept of ``traffic'' and remove any subjectivity in interpretation, we must clearly define it within the context of our prompts. Specifically, we refer to traffic as any situation where external factors are causing the vehicle to reduce speed. We operationalize this in the model prompts with the following explicit query: ``Is there traffic causing the car to slow down?'' (as illustrated in Fig.~\ref{fig:vid-setup}). This precise definition ensures consistency across evaluations and reduces ambiguity. When testing, we observed that models consistently yielded higher accuracy rates in identifying traffic scenarios compared to cases focused on ego-motion, as reflected in Table~\ref{tab:accuracy}. This case also presents a clear example of how GPT-4o has improved upon the prior model, achieving 77\% accuracy and thus, a higher level of reasoning. We explore further results, notably showcasing the changes from GPT-4V (March) to the latest GPT-4o, presented in Fig.~\ref{fig:traffic-plot}, illustrating the models' performance using a confusion matrix that captures predictions across various traffic levels. 
Overall, The prediction positively correlates to traffic level where Claude3 is more successful at identifying high levels while GPT-4V (March) is particularly successful at identifying a lack of traffic. We see the marked improvement in identifying traffic in GPT-4o, able to shake what may have been a \textit{no traffic} bias in the older model. While not perfect, the models' ability to identify traffic was the highest level of success achieved across ego-motion and other actor scenarios.

\begin{figure}[t]
 \centering
 \includegraphics[width=0.5\textwidth]{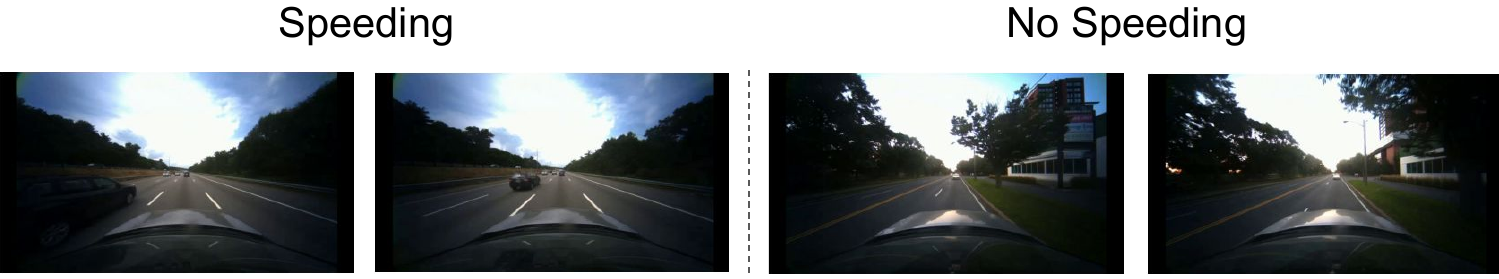}
   \vspace{-2mm}
 \caption{Speeding/no speeding frames from real-footage scene}\label{fig:speeding-frames}
\end{figure}

\noindent\textbf{Speeding vehicle vs no speeding vehicle} identification is critical for road safety. This requires an understanding of geometry and time for perceiving the motion of another actor. Here, we provide scenarios from real-footage where the ego vehicle is overtaken by another vehicle or not and from \METHOD{}, involving another actor that is speeding or not speeding at two-speed levels each. To emphasize this point, we must clarify that the ego vehicle is driving at the speed limit such that the relative speed of the other agent can be observed accurately. Given the aforementioned formulation, a large change in distance between the ego car and the other vehicle should indicate speeding and be sufficient for a human to understand the presence of a speeding vehicle. 
This philosophy is applied to the real footage as well, where a ``speeding'' vehicle is present when another vehicle overtakes the ego vehicle, and given the ego vehicle is clarified to be driving at the speed limit, the overtaking vehicle is implied to be speeding, as shown in Fig.~\ref{fig:speeding-frames}.

\begin{figure}[t]
    \centering
    \includegraphics[width=0.5\textwidth]{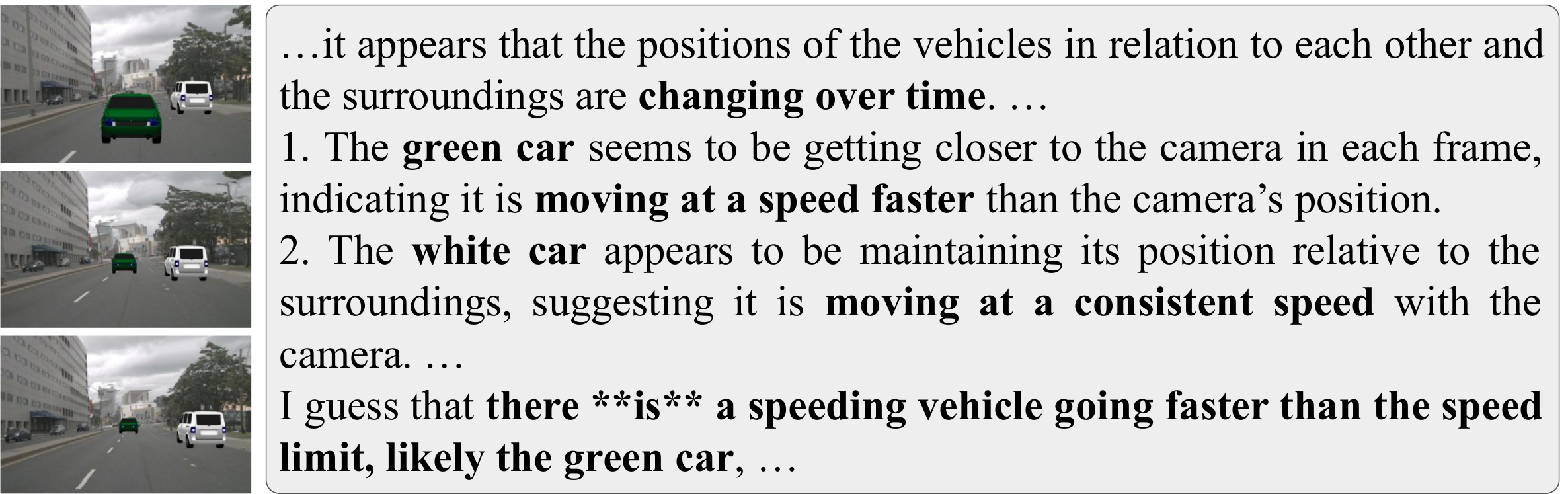}
    \caption{Frames from a speeding scenario with an added vehicle traversing at normal speed provided by \METHOD{} with the GPT-4V response.}
    \label{fig:special-speeding-frames}
\end{figure}

As previously, we must remove ambiguity in the query, adjusting the prompt shown in Fig.~\ref{fig:vid-setup} to ensure the model is aware that the ego vehicle is driving at the speed limit. Therefore, we make the following key change: ``All are taken from the same camera that is fixed on a moving car going at the speed limit''. When observing the results in Table \ref{tab:accuracy}, we see a roughly 50\% accuracy in the models. We further analyze these results, looking at the percentage of responses from the models that were either \textit{speeding} or \textit{no speeding} regardless of the ground truth. Despite the contrast in the motions, 
GPT-4V barely detected a speeding vehicle, reporting there was no speeding vehicle 94\% of the time across real and simulated scenes, a trend shared by the smaller models. Claude3 has a strong bias towards reporting speeding at 87.5\% of all scenes. This level of extreme bias was not present in GPT-4o, where instead the model is the clear front runner in detecting a speeding vehicle, and in other actor behavior reasoning in general. 
This is an interesting case of biases that requires further probing. 
With the functionality of \METHOD{}, we can augment one of the many GPT-4V failed scenarios as shown in Fig.~\ref{fig:special-speeding-frames}. As such, we introduce a reference vehicle traversing at the same speed as the ego-car to the scene while providing the same text prompt which yielded a positive result for GPT-4V, as it successfully identified the speeding vehicle. As such, the bias is not likely due to biases in the expected motions of vehicles (as with the forward bias) but more likely due to failure to understand the relation of the ego-car to its surrounding world, necessitating a very clear reference point, beyond one that would be present in real footage, whereas GPT-4o seemingly had better capabilities in identifying a reference point and not suffer this limitation.


\subsection{Open-Set Reasoning}
\begin{figure}[t]
    \centering
        \includegraphics[width=\linewidth]{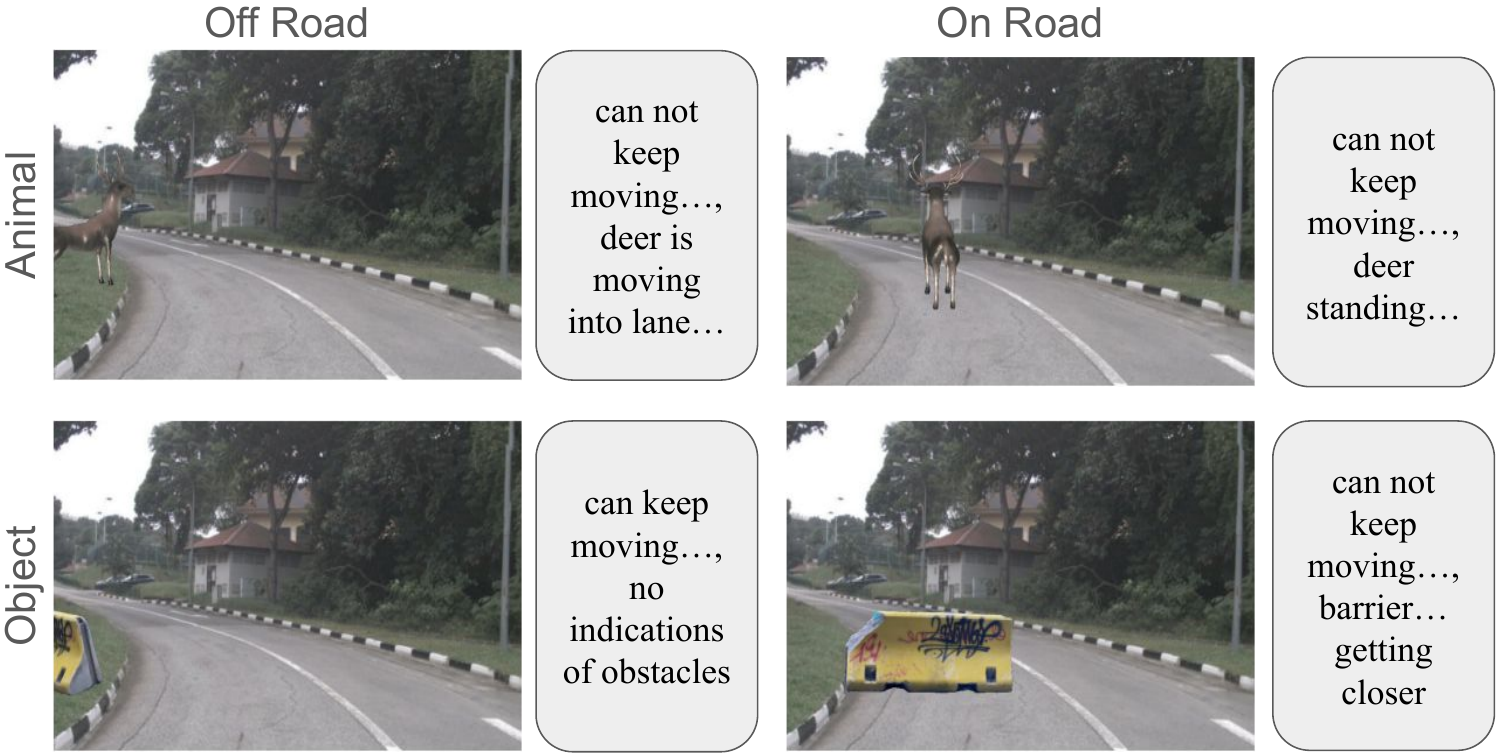}
        \caption{Frames from object and animal scenarios with truncated GPT-4o responses.}
        \label{fig:object-frames}
\end{figure}

The seemingly random placement of \textbf{animals and static objects} in a scene is one of the open-set scenarios \METHOD{} enables to evaluate MLLMs. 
As human drivers, we handle these situations instinctively: slow down or avoid static objects on the road, while ignoring those off-road. For animals, uncertainty leads to slowing down or avoiding them regardless. The actions are clear in the scenarios shown in Fig~\ref{fig:object-frames}. 
We can see that the large models, GPT-4o, GPT-4V, and Claude3, were quite successful in their reasoning for these cases as shown in Table \ref{tab:accuracy} and as seen by GPT-4o responses in Fig.~\ref{fig:object-frames}; however, note that the accuracy of GPT-4o is actually lower than GPT-4V Legacy model so in this case, the newer model has not achieved an improvement.

\begin{figure}[t]
        \centering
        \includegraphics[width=\linewidth]{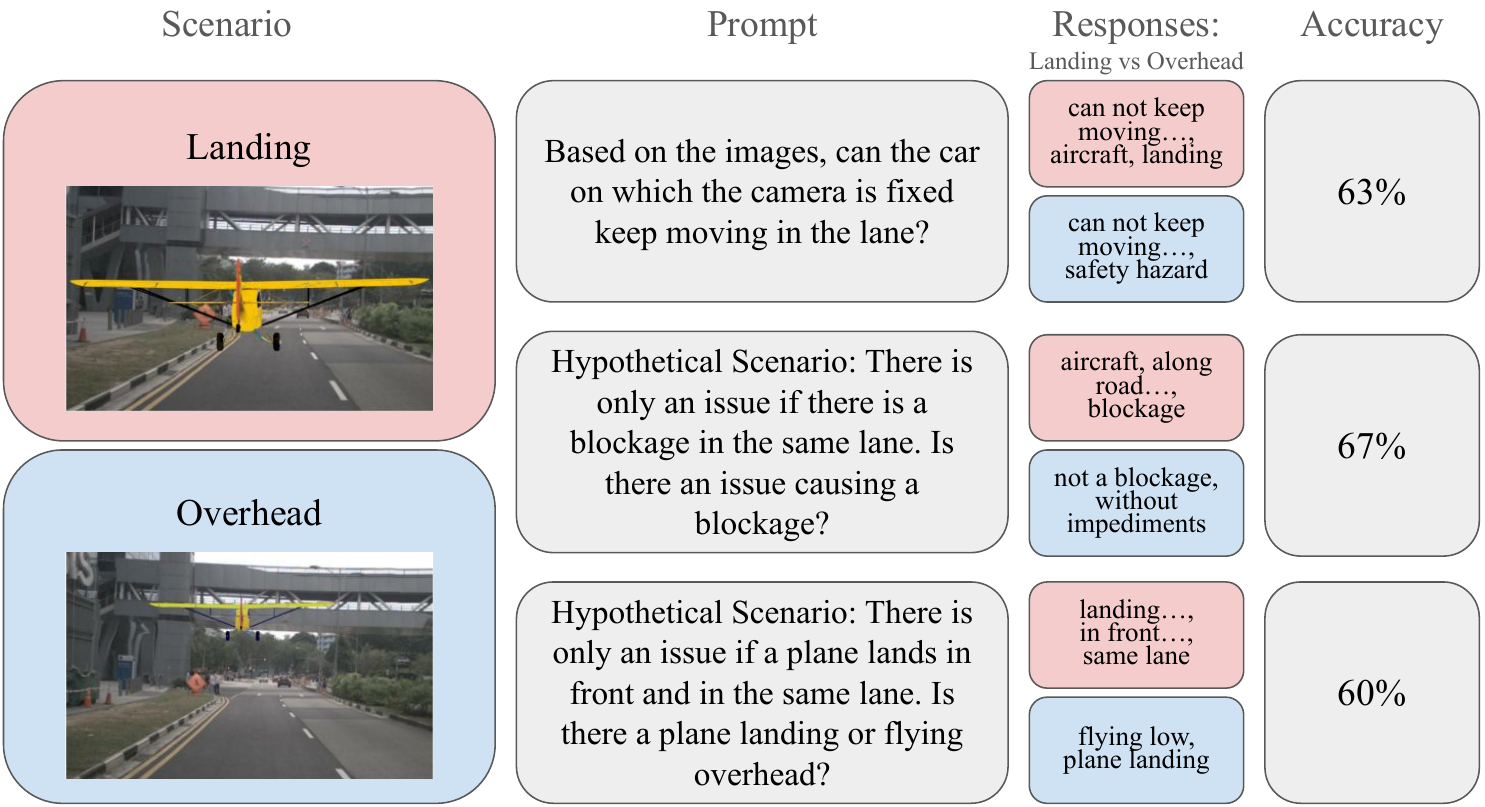}
        \caption{Plane scenarios with truncated prompts and GPT-4V Legacy responses with accuracy.}
        \label{fig:plane-frames-prompts-responses}
\end{figure}

\noindent\textbf{Plane landings vs flying overhead} is a fascinating open-set scenario we explore with \METHOD. A human driver may not know how to react to such an extreme case but we can observe the MLLMs behavior. The frames in Fig.~\ref{fig:plane-frames-prompts-responses} showcase a scene with the plane landing or flying overhead. The primary prompt, used for Table \ref{tab:accuracy}, showed that regardless of the plane landing, the model suggested that you can not keep moving due to the risk: a fair response. As such, we explored a few hypothetical scenarios which really test the geometric and temporal understanding of the plane's motion, which is in a completely different axis than other scenarios.

\subsection{Planning Reasoning}
\begin{figure}[t]
        \centering
        \includegraphics[width=\linewidth]{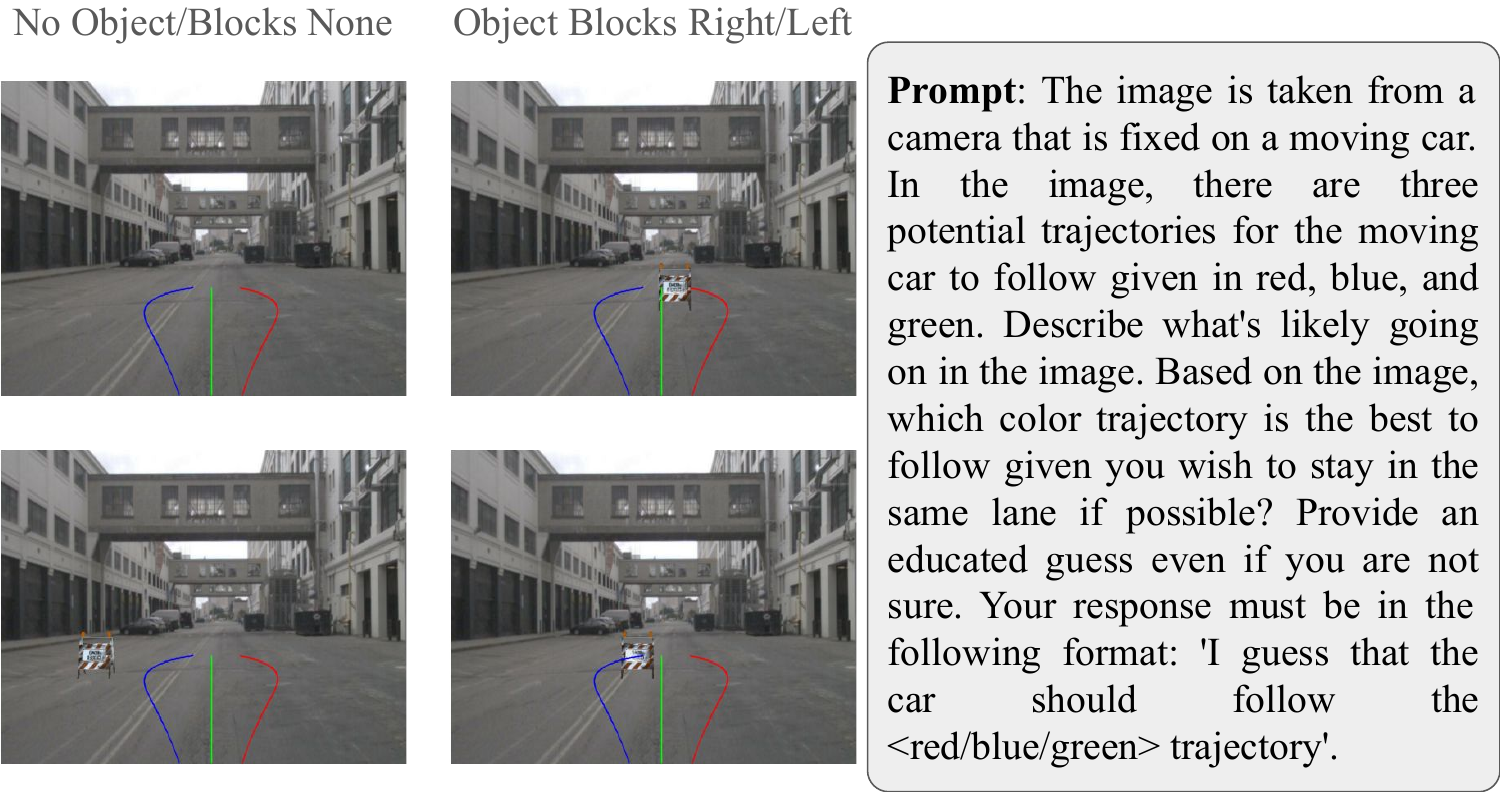}
        \caption{Images depicting plans with different static object placements and the text prompt.}
        \label{fig:planning-frames}
\end{figure}

We present planning experiments using \METHOD, generating plans for map points and visualizing them in the camera view. We also introduce static objects to assess whether MLLMs can choose paths that navigate around obstacles. In Fig.~\ref{fig:planning-frames} we show the four ways we ran planning evaluations for a given scene: (1) no object, (2) the object not blocking anything, (3) the object blocking the middle and right trajectories, and (4) the object blocking the middle and left trajectories. Given the aim of staying in the same lane and presented with the three trajectory choices, a human driver would have a clear pick for each example: (1) green, (2) green, (3) blue, and (4) red.  To run the evaluation, we utilize a different style of prompt to pick a trajectory in a single image. As such, we used the prompt shown in Fig.~\ref{fig:planning-frames}, where we also specify the objective of staying in the same lane so there is always one correct choice.
We see in Table~\ref{tab:accuracy} that the larger models, GPT-4o, GPT-4V, and Claude3, achieve significantly better accuracy than the others. However, \textbf{their success rate is still surprisingly, under $50\%$,} which is not ideal 
for closed-loop planning.
As such, further probing is required to see the source of the limitations. We improved Claude3's performance from 45\% to 55\% by adding ``while avoiding obstacles'' to the prompt. This addition showcases the failure of the MLLM as a world model to boost accuracy. 

\subsection{Real versus Simulated Data}
Table \ref{tab:accuracy} shows similar results between real and simulated data, with a few exceptions, like Claude3 in \textit{traffic} tests. However, the model's response patterns remained consistent across data types, as seen in in Table \ref{tab:forwardbackward}, and those we discussed when exploring the \textit{speeding} results. Looking at the underlying response strategy for Claude3 in a similar manner to those shown in Table \ref{tab:forwardbackward}, we see on real data, the responses for \textit{traffic} and \textit{no traffic} are both 50\% where for simulated, the responses were 55\% and 45\% respectively, so the model was effectively at a random guess level (this can also be seen with the lack of monotonic behavior for Claude3 in Fig.~\ref{fig:traffic-plot}). Overall, the average absolute difference between real versus sim across all models is $0.05$, and the typical behavior observed in their response strategy is consistent.

\section{CONCLUSION}
This work demonstrates the current capabilities of SOTA MLLMs, including GPT-4o and Claude3, as driving world models. Despite 
the improvements in GPT-4o over the prior best GPT model, their limitations in reasoning across multiple frames of driving scenarios have become evident through our extensive experimental results. While many accuracy levels seem random, \METHOD{} allows probing the reasoning capabilities that are behind the prediction, exposing details on biases. We observed that failures in handling scenarios stemmed from biases in expected vehicle movements, like assuming forward motion on a road.
\addtolength{\textheight}{0cm}   






\bibliographystyle{IEEEtran}
\bibliography{IEEEabrv,root}




\end{document}